\pdfoutput=1

\documentclass[letterpaper, 10 pt, conference]{ieeeconf}  

\IEEEoverridecommandlockouts                              
\let\labelindent\relax

\usepackage{graphicx} 
\usepackage{epsfig} 
\usepackage{amsmath} 
\usepackage{amssymb} 
\usepackage{pifont}
\usepackage{soul,color} 
\usepackage{mathtools}
\usepackage{booktabs}
\usepackage{multirow}
\usepackage{enumerate}
\usepackage{balance}
\usepackage{gensymb}
\usepackage{tabulary}
\usepackage[numbers,sort&compress]{natbib}
\usepackage{dblfloatfix}
\usepackage{layouts}
\usepackage{algorithm}
\usepackage{algpseudocode}
\usepackage{lipsum}
\usepackage{subcaption}
\usepackage{enumitem}
\usepackage{siunitx}
\usepackage[bookmarks=false]{hyperref}
\usepackage{float}

\graphicspath{{figures/}}

\usepackage{subcaption}
\usepackage{tabulary}
\usepackage{multirow}

\usepackage{mwe}
\usepackage{xcolor}
\begin{document}

\title{\LARGE \bf
Ground Encoding: Learned Factor Graph-based Models for Localizing Ground Penetrating Radar
}

\author{Alexander Baikovitz$^{1}$, Paloma Sodhi$^{1}$, Michael Dille$^{2}$, Michael Kaess$^{1}$
\thanks{This work was supported by a NASA Space Technology Graduate Research Opportunity.}
\thanks{$^{1}$ The Robotics Institute, Carnegie Mellon University, $^{2}$ Intelligent Robotics Group, NASA Ames Research Center, Correspondence to 
        {\tt\small abaikovitz@cmu.edu}}}%

\maketitle
\thispagestyle{empty}
\pagestyle{empty}

\begin{abstract}

We address the problem of robot localization using ground penetrating radar (GPR) sensors. Current approaches for localization with GPR sensors require \textit{a priori} maps of the system's environment as well as access to approximate global positioning (GPS) during operation. In this paper, we propose a novel, real-time GPR-based localization system for unknown and GPS-denied environments. We model the localization problem as an inference over a factor graph. Our approach combines 1D single-channel GPR measurements to form 2D image submaps. To use these GPR images in the graph, we need sensor models that can map noisy, high-dimensional image measurements into the state space. These are challenging to obtain \textit{a priori} since image generation has a complex dependency on subsurface composition and radar physics, which itself varies with sensors and variations in subsurface electromagnetic properties. Our key idea is to instead learn relative sensor models directly from GPR data that map non-sequential GPR image pairs to relative robot motion. These models are incorporated as factors within the factor graph with relative motion predictions correcting for accumulated drift in the position estimates. We demonstrate our approach over datasets collected across multiple locations using a custom designed experimental rig. We show reliable, real-time localization using only GPR and odometry measurements for varying trajectories in three distinct GPS-denied environments.

\end{abstract}

\section{Introduction}

We focus on the problem of localization using ground penetrating radar (GPR) in unknown and GPS-denied environments. Hardware failure, repetitive or sparse features, and poor visibility and illumination can make localization in warehouses, mines, caves and other enclosed, unstructured environments challenging. Consider an operational subsurface mine where continuous drilling and blasting changes the line-of-sight appearance of the scene and creates unexplored environments, which could lead to poor localization using visual and spatial information. While the visual environment may change, subsurface features are typically invariant and can be used to recognize the system's location.

\begin{figure}[!t]
    \centering
    \includegraphics[width=\columnwidth]{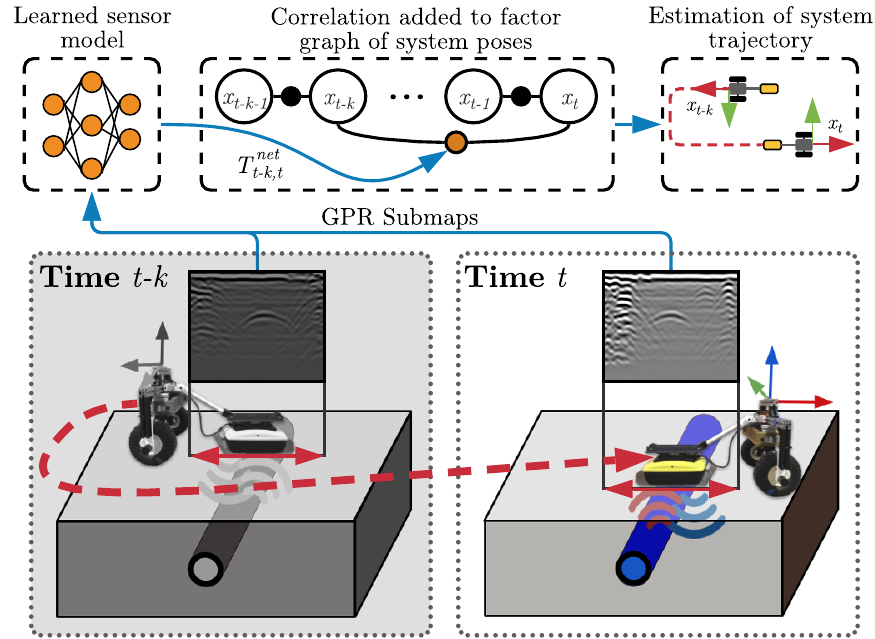}
    \caption{Estimating poses for a ground vehicle using subsurface measurements from Ground Penetrating Radar (GPR) as inference over a factor graph. Since GPR measurements are challenging to correlate, a relative transformation is learned between submaps to correct the system's pose.}
    \label{fig:cover}
\end{figure}

Currently, GPR is widely used in utility locating, concrete inspection, archaeology, unexploded ordinance and landmine identification, among a growing list of applications to determine the depth and position of subsurface assets~\cite{JOL2009xiii,PajewskiGPRapplication}. GPR information is often difficult to interpret because of noise, variable subsurface electromagnetic properties, and sensor variability over time~\cite{ernenwein2008}. A single GPR measurement only reveals limited information about a system's environment, requiring a sequence of measurements to discern the local structure of the subsurface. In order to use GPR for localization, we need a representation for GPR data that (1) captures prominent local features and (2) is invariant over time.

In GPS-denied environments, operating using only proprioceptive information (e.g. IMU, wheel encoders) will accumulate drift. Prior work~\cite{Cornick2016,Ort2020} on GPR localization has addressed this challenge by using an \textit{a priori} global map of arrayed GPR images. During operation, the system compares the current GPR image against this global map database, and corrects for accumulated drift within a filtering framework. However, this approach requires approximate global positioning and a prior map of the operating environment which would not generalize to previously unknown environments. 

We propose an approach that allows for correction of this accumulated drift without any prior map information. In the absence of a prior map, we must reason over multiple GPR measurements together to be able to infer the latent robot location. We formulate this inference problem using a factor graph, which is now common with many modern localization and SLAM objectives~\cite{dellaert2017factor,Cadena2016,czarnowski2020deepfactors,sodhi2020learning,Bloesch2018CodeSlam}. GPR, inertial, and wheel encoder measurements are incorporated into the graph as factors to estimate the system's latent state, consisting of position, orientation, velocity, and IMU biases. To incorporate measurements into the graph, a sensor model is needed to map measurements to states. For GPR sensors, \textit{a priori} models are typically challenging to obtain since image generation has a complex dependency on subsurface composition and radar physics. Instead, we learn relative sensor models that map non-sequential GPR image pairs to relative robot motion. The relative motion information in turn enables us to correct for drift accumulated when using just proprioceptive sensor information. Our main contributions are:
\begin{enumerate}
    \item A formulation of the GPR localization problem as inference over a factor graph without a prior map.
    \item A learnable GPR sensor model based on submaps.
    \item Experimental evaluation on a test platform with a single-channel GPR system operating in three different GPS-denied environments.
\end{enumerate}

\section{Related Work}
\textbf{GPR in robotics:} Most use of GPR in the robotics domain has been for passive inspection, which includes excavation of subsurface objects~\cite{Herman1997}, planetary exploration~\cite{furgale2010,hamran2015, Lai2020}, mine detection~\cite{Nonami2009, Sato2009}, bridge deck inspection~\cite{La2013Bridge}, and crevasse identification~\cite{6858016}. 

Localizing GPR (LGPR) was introduced by Cornick \textit{et al.} to position a custom GPR array in a prior map~\cite{Cornick2016}. In this method, the current GPR measurement is registered to a grid of neighboring prior map measurements using a particle swarm optimization, which attempts to find the 5-DOF position of the measurement that maximizes the correlation with the grid data. Ort \textit{et al.} extended this work using the same LGPR device for fully autonomous navigation without visual features in variable weather conditions~\cite{Ort2020}. This approach involves two Extended Kalman Filters; one filter estimates the system velocities from wheel encoder and inertial data and the other filter fuses these velocities with the LGPR-GPS correction. In the evaluation across different weather conditions, it is apparent that GPR measurements acquired in rain and snow were less correlated to the \textit{a priori} map than in clear weather using the heuristic correlation for registration method proposed in~\cite{Cornick2016}.

\textbf{Factor graphs for SLAM:} Modern visual and spatial localization systems often use smoothing methods, which have been proven to be more accurate and efficient than classical filtering-based methods~\cite{Cadena2016}. The smoothing objective is typically framed as a MAP inference over a factor graph, where the variable nodes represent latent states and factor nodes represent measurement likelihoods. To incorporate sensor measurements into the graph, a sensor model is needed to map high dimensional measurements to a low-dimensional state. Typically these sensor models are analytic functions, such as camera geometry~\cite{Mur-Artal2018, Engel2014} and scan matching~\cite{Zhang2014LOAM}.

\textbf{Learned sensor models:} Learning-based methods provide an alternate option to model complex sensor measurements. Prior work in visual SLAM has produced dense depth reconstructions from learned feature representations of monocular camera images~\cite{czarnowski2020deepfactors, Bloesch2018CodeSlam}. In computer vision, spatial correlation networks have been used to learn optical flow and localize RGB cameras in depth maps~\cite{fischer2015flownet, cattaneo2019cmrnet, cattaneo2020cmrnet}. Recent work on object state estimation using tactile feedback has demonstrated the effectiveness of learned sensor models in factor graph inference~\cite{sodhi2020learning}.

Our approach enables robust GPR-based positioning by leveraging the benefits of structured prediction in factor graph inference with learning-based sensor models. We use a smoothing-based approach, which is more accurate and efficient than filtering-based approaches used by previous localizing GPR systems, and can be solved in real-time using state-of-the-art algorithms~\cite{kaess2008isam, kaess2012isam2}. Learning-based sensor models for GPR can outperform engineered correlation approaches by exploiting unique features in a sequence of GPR measurements, enabling improved performance in even sparsely featured environments.

\section{Problem Formulation}
\label{sec:probform}

We formulate our GPR localization problem as inference over a factor graph. A factor graph is a bipartite graph with two types of nodes: variables $x\in\mathcal{X}$ and factors $\phi(\cdot): \mathcal{X}\rightarrow \mathbb{R}$. Variable nodes are the latent states to be estimated, and factor nodes encode constraints on these variables such as measurement likelihood functions. 

Maximum a posteriori (MAP) inference over a factor graph involves maximizing the product of all factor graph potentials, i.e.,
\begin{equation}
	\begin{split}
	\label{eq:probform:eq3.1}
	\hat{x}^{}& = \underset{{x}}{\operatorname{argmax}}\ \prod_{i=1}^{m} \phi_i({x}) \\
	\end{split}
\end{equation}

Under Gaussian noise model assumptions, MAP inference is equivalent to solving a nonlinear least-squares problem~\cite{dellaert2017factor}. 
That is, for Gaussian factors $\phi_i({x})$ corrupted by zero-mean, normally distributed noise,
\begin{equation}
	\begin{split}
		\label{eq:probform:eq3.2}
    	&\phi_i({x})\propto \exp\left\{-\frac{1}{2}||f_i({x})-z_i||_{\Sigma_i}^2\right\} \\
		\Rightarrow \ & \hat{x}^{}=\underset{{x}}{\operatorname{argmin}}\sum_{t=1}^{T}||f_i({x})-z_i||_{\Sigma_i}^2 \\
	\end{split}
\end{equation}
where, $f_i({x})$ is the measurement likelihood function predicting expected measurement given current state, $z_i$ is the actual measurement, and $||\cdot||_{\Sigma_i}$ is the Mahalanobis distance with measurement covariance $\Sigma_i$.

For the GPR localization problem, variables in the graph at time step $t$ are the 6-DOF robot poses $s_t \in SE(3)$, velocities $v_t$, and IMU bias $b_{t}$, i.e. $x_t=[s_t\ v_t\ b_t]^T$. Factors in the graph incorporate different likelihoods for GPR, IMU, and wheel encoder measurements. At every time step $t$, new variables and factors are added to the graph. Writing out Eq. \ref{eq:probform:eq3.2} for the GPR localization objective,
\begin{equation}
	\begin{split}
		\label{eq:probform:eq3.3}
		\hat{x}_{1:T} = & \underset{x_{1:T}}{\operatorname{argmin}}\sum_{t=1}^{T} \left\{||f_{gpr}({x_{t\text{-}k}, x_t})-z_{t\text{-}k,t}^{gpr}||^{2}_{\Sigma_{gpr}}+\right. \\ 
		& \left. ||f_{wh} ({x_{t\text{-}1}, x_t})-z_{t\text{-}1,t}^{wh}||^{2}_{\Sigma_{wh}} +\right. \\
		& \left. ||f_{imu}({x_{t\text{-}1}, x_t})-z_{t\text{-}1,t}^{imu}||^{2}_{\Sigma_{imu}} \right\}
	\end{split}
\end{equation}
This objective in Eq. \ref{eq:probform:eq3.3} is solved online every time step using an efficient, incremental solver iSAM2~\cite{kaess2012isam2}. 
Individual terms in Eq. \ref{eq:probform:eq3.3} are described in more detail in Section \ref{sec:approach}.

\section{Approach}
\label{sec:approach}

\begin{figure*}[!t]
    \centering
    \includegraphics{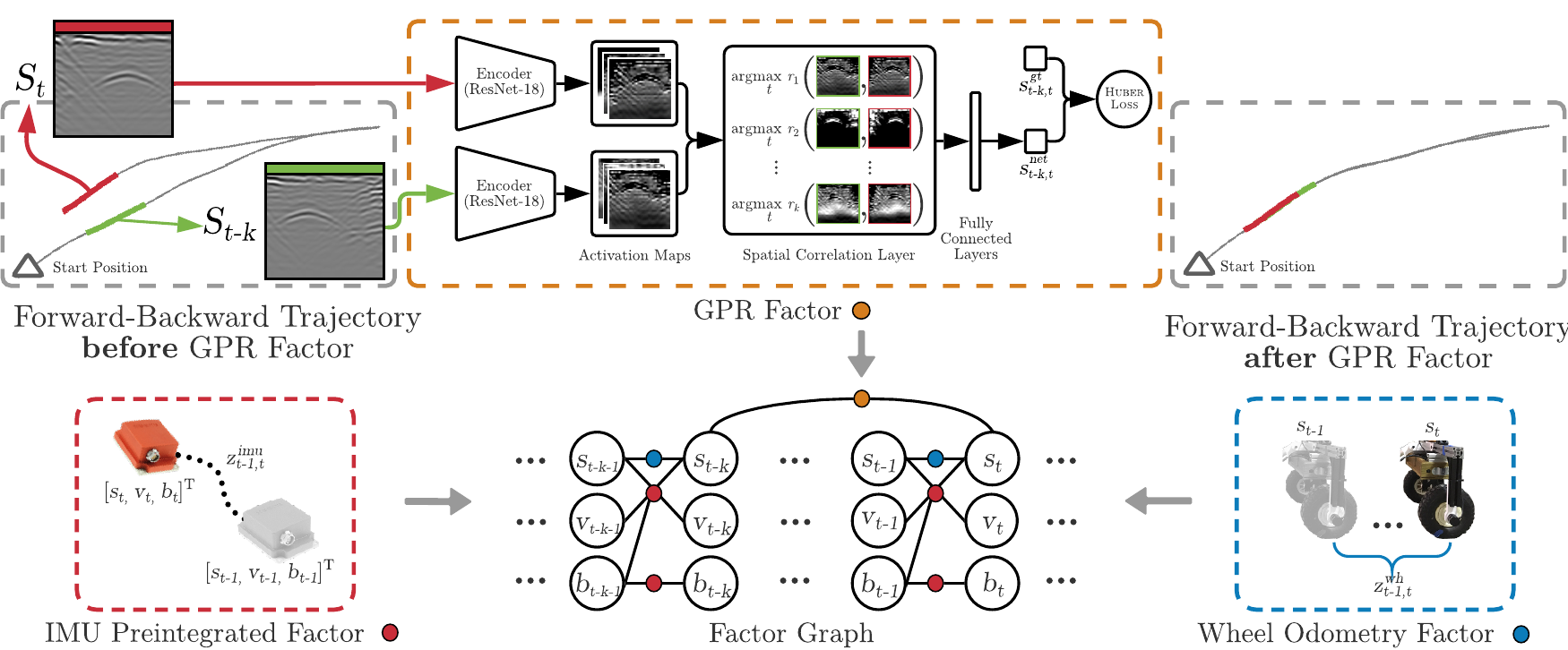}
    \caption{Overview of the Ground Encoding system. On the left, two submaps containing similar subsurface structures are acquired during forward-backward trajectory that has accumulated drift. An autoencoder maps these submaps to feature activation maps, which are used by the transform prediction network to find the relative transformation. Factors denoting the measurement likelihoods from GPR, inertial, and wheel encoder measurements are incorporated into the graph to estimate the system's latent states $x_t = [s_t, v_t, b_t]^T\ \forall t \in \{1\ ...\ T\}$. The trajectory on the right shows the outcome of the GPR correction.}
    \label{fig:pipeline_overview}
\end{figure*}

There are three primary stages in our approach: GPR pre-processing and submapping (Section \ref{sec:approach:submapping}), learning GPR models (Section \ref{sec:approach:senor_model}), and factor graph optimization (Section \ref{sec:approach:factorgraph}). Fig.~\ref{fig:pipeline_overview} illustrates our ground encoding approach.  

\subsection{GPR Submapping}
\label{sec:approach:submapping}

\begin{figure}
    \centering
    \includegraphics{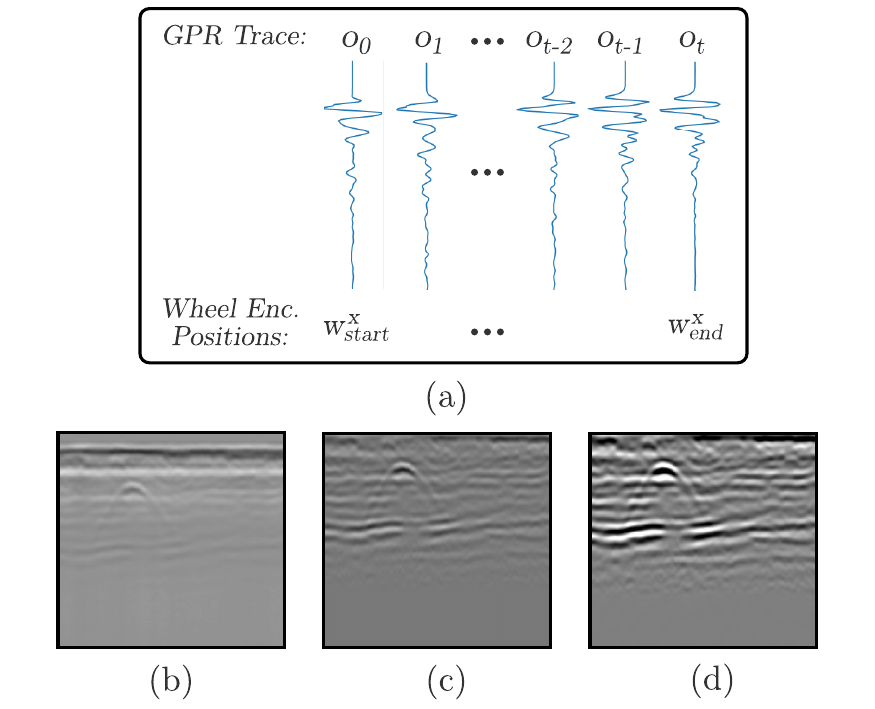}
    \caption{Overview of GPR image construction. \textbf{(a)} Unprocessed localized traces received by the device. \textbf{(b)} Horizontally stacked traces from (a), where the amplitudes correspond to pixel intensity. \textbf{(c)} Measurements after filtering and gain. \textbf{(d)} Final image after thresholding.}
    \label{fig:gpr_images}
\end{figure}
\textbf{Preprocessing:} Digital signal processing techniques are needed to improve the signal-to-noise ratio of the received GPR signal, counteract signal attenuation, and remove recurrent artifacts that are intrinsic to GPR systems. Measurements $o_t$ arrive from each position as a 1D discrete waveform called a trace, which are locally averaged to reduce noise and resampled to form a uniformly spaced image~\cite{4671101}.

Receiving signals often contain a low frequency component and DC bias caused by saturation and inductive coupling, requiring a dewow filter, which involves a DC subtraction and low-cut filter parameterized on the radar's center frequency and bandwidth~\cite{JOL2009xiii}. In order to counteract attenuation, we multiply the signal by a Spreading and Exponential Compensation (SEC) gain function,
\begin{equation}
    G = \exp(a \cdot t) \cdot t^b
\end{equation}
where, $a$ is the exponential gain constant and $b$ is the power gain constant~\cite{huber2018}.

GPR images have global and local context. Global features such as horizontal layers encode information about the general environment of a system, which can include boundaries like layers of asphalt in roads. Local features, which include pipes and point objects, provide salient features needed to effectively localize. To maintain prominent global features, we subtract the mean trace over all prior submaps (as opposed to over each submap) to remove repeated artifacts introduced by the radar while emphasizing local features.

\textbf{Submap formation:}
Traditional GPR systems provide a one-dimensional trace ${o}_t$ at each position, which does not provide enough information to effectively determine a system's unique position. A collection of traces represents a local fingerprint that may contain valuable information for localization. Our approach involves constructing submaps $S_t$ based on integration of local wheel encoder measurements ${w}_t^x$ (Fig. \ref{fig:gpr_images}). A submap is approximated by sampling discrete GPR measurements from a continuous distribution of GPR measurements using an interpolation function $\hat{F}$ to create a uniformly sampled image,
\begin{equation}
    S_t = \hat{F}(\{(\bar{o}_t, {w}_{t}^x) \mid {w}_{start}^x \le {w}_{t}^x < {w}_{end}^x\})
\end{equation}
where, $\bar{o}_t$ is the locally averaged GPR measurement, described in Preprocessing, at time $t$ and $w_{t}^x$ is the wheel encoder measurement at time $t$. We use a rule-based method that maps a set of wheel odometry and angular velocity measurements to a decision of whether a short trajectory is a valid submap.

\subsection{Learned GPR model}
\label{sec:approach:senor_model}
The goal of the learned GPR sensor model is to derive the GPR factor cost term in Eq. \ref{eq:probform:eq3.3} for the factor graph optimization. The sensor model is trained to predict a relative transformation between two submaps $\{S_{t-k}, S_t\}$ from ground truth measurements. Submap images are compressed into feature activation map, which are provided to a transform prediction network to estimate the relative transformation $s_{t-k,t}^{net}$ between poses $s_{t-k}$ and $s_{t}$ (Fig. \ref{fig:pipeline_overview}).

\textbf{Feature learning:} \label{sec:feat_learning}
Identifying the function that directly maps submaps to transformations is prone to overfitting because of substantial noise in the original radar submaps. To learn the function that relates two similar submaps $S_{t-k}$ and $S_{t}$, an intermediate feature representation $f_{t-k}$ and $f_t$ is needed. A ResNet-18 auto-encoder is used to obtain $k$ feature activation maps that contain relevant features for localization like vertical edges. The autoencoder is trained with an $L1$ reconstruction loss to preserve the sparsity of features in the original submap.

\textbf{Submap comparison:}
Prior to identifying transformations in the data, we perform an average pooling on the row-wise standard deviation of submap images to check for salient features. If submaps $S_{t-k}$ and $S_{t}$ contain valid features, a linear correlation network checks whether they share common features using a method further described in Transform Prediction.

\textbf{Transform prediction:}
If submaps contain salient features and are correlated, we predict a relative 1D transformation $s^{net}_{t-k,t}$ using a two stage approach. We first construct a set of cost curves by comparing each feature map $S_{t-k,i}$ and $S_{t,i}$ for all $k$ using Eq.~\ref{eq:approach:analytic:corr} for feature activation map pixel space. We then construct a vector of $\operatorname{argmax}$ elements from each cost curve, which is provided to a fully-connected regression network to identify the relative transformation $s^{net}_{t-k,t}$. The network learns to provide higher weight to filters that are better correlated with the desired translation, reducing the effect of filters that encode common patterns in the data like horizontal banding. The transform prediction network is trained using Huber loss against supervised ground truth data $s^{gt}_{t-k,t}$ acquired by a robotic total station.

\subsection{Factor graph optimization}
\label{sec:approach:factorgraph}

\textbf{GPR factor:} Measurements from GPR are incorporated into the graph as factors with the cost from Eq.~\ref{eq:probform:eq3.3}. The relative GPR factor has a quadratic cost penalizing large residual values and is defined as,
\begin{equation}
	\begin{split}
		\label{eq:approach:gpr:eq1}
		||f_{gpr}({x_{t\text{-}k}, x_t})-z_{t\text{-}k,t}||^{2}_{\Sigma_{gpr}}\ :=\ ||s^{-1}_{t}s_{t\text{-}k}\ominus z_{t\text{-}k, t}^{gpr}||^{2}_{\Sigma_{gpr}}
	\end{split}
\end{equation}
where, $z_{t\text{-}k,t}^{gpr}$ is the predicted relative transformation from the transform prediction network, $s^{-1}_{t}s_{t\text{-}k}$ is the estimated relative pose transformation between two variable nodes in the graph, and $\ominus$ represents the difference between two manifold elements.

\textbf{IMU preintegrated factor:}
For incorporating IMU measurements in the graph, we use the IMU preintegration factor proposed in~\cite{Forster}. The preintegration factor locally integrates multiple high-frequency accelerometer and gyroscope measurements into a single compound preintegrated IMU measurement. This has the advantage of combining the speed and complexity benefits of filtering-based methods along with the accuracy of smoothing methods.

The IMU factor term from Eq. \ref{eq:probform:eq3.3} can be expressed as a residual over the differences in orientation $\Delta R_{i,j}$, velocity $\Delta v_{i,j}$, and position $\Delta t_{i,j}$,
\begin{equation}
\begin{split}
\label{eq:approach:preint:eqn0}
    ||f_{imu}({x_{t\text{-}1}, x_t}) & -z_{t\text{-}1,t}||^{2}_{\Sigma_{imu}} = ||\mathbf{r}_{\mathcal{I}_{ij}}||^{2} + ||\mathbf{r}_{b_{ij}}||^{2}
\end{split}
\end{equation}
where, $\{i,j\}$ are state index pairs between which preintergration is performed. We perform the preintergration between consecutive states, i.e. $\{i,j\}:=\{t\text{-}1, t\}$. 
Here, $\mathbf{r}_{\mathcal{I}_{ij}}=[\mathbf{r}_{\Delta R_{i,j}}^T, \mathbf{r}_{\Delta v_{i,j}}^T, \mathbf{r}_{\Delta t_{i,j}}^T]^T$ is the preintegration error residual and $\mathbf{r}_{b_{ij}}$ is bias term estimation errors. We refer the reader to \cite{Forster} for more details on these residual error terms.

\textbf{Wheel encoder factor:} Wheel encoder measurements are incorporated into the graph between sequential poses $\{s_{t-1}, s_{t} \}$. The relative wheel encoder factor is defined as,
\begin{equation}
	\begin{split}
		\label{eq:approach:wheel_encoder:eqn1}
		||f_{wh}({x_{t\text{-}1}, x_t})-z_{t\text{-}1,t}^{wh}||^{2}_{\Sigma_{wh}} := ||s^{-1}_{t}s_{t\text{-}1}\ominus z_{t, t\text{-}1}^{wh}||^{2}_{\Sigma_{wh}}
	\end{split}
\end{equation}
where, $z_{t\text{-}k,t}^{wh}$ is the relative difference between two poses measured by the wheel encoder.

\section{Results and Evaluation}

We evaluate our GPR-based localization system in three distinct, GPS-denied environments: a basement (\textit{nsh\_b}), a factory floor (\textit{nsh\_h}), and a parking garage (\textit{gates\_g}). We first discuss our experimental setup in Section~\ref{sec:eval:setup}. We then provide an ablation of validation losses from different learned and engineered GPR sensor models in Section~\ref{sec:eval:learning}. We finally evaluate our entire GPR-based localization system and demonstrate the effects of adding GPR information in the graph optimization in Section~\ref{sec:eval:opt}.

\subsection{Experimental setup}
\label{sec:eval:setup}
\begin{figure}[!t]
    \centering
    \includegraphics{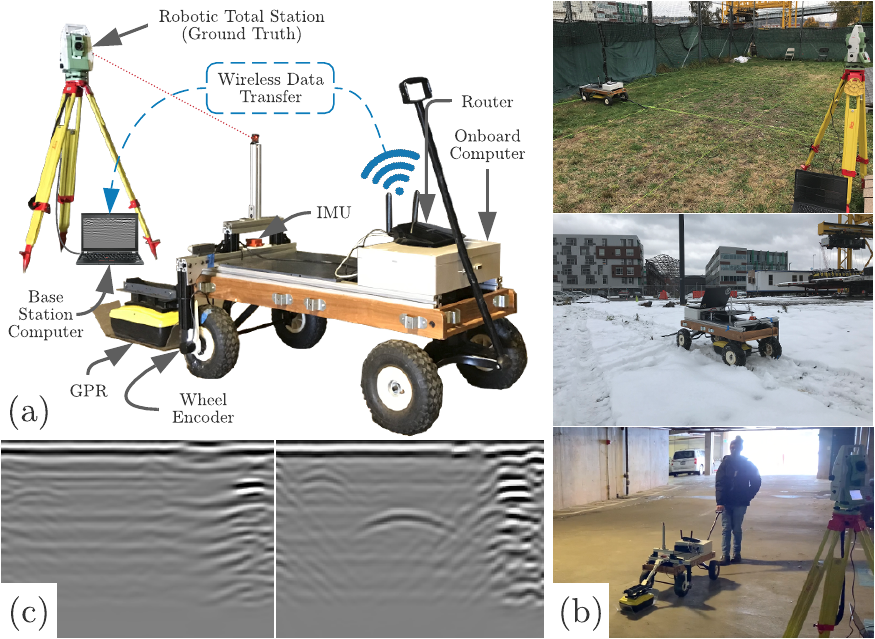}
    \caption{\textbf{(a)} Experimental setup for ground encoding datasets. \textbf{(b)} Testing in distinct environments. \textbf{(c)} Left GPR image constructed with 2.5cm antenna-ground separation. Right GPR image constructed with ground-coupled antenna configuration.}
    \label{fig:exp_setup}\vspace{-3mm}
\end{figure}

We constructed a manually-pulled test rig named SuperVision shown in Fig.~\ref{fig:exp_setup}(a) for data acquisition. SuperVision uses a quad-core Intel NUC for compute and wirelessly transmits onboard data from an XSENS MTI-30 9-axis Inertial Measurement Unit, YUMO quadrature encoder with 1024 PPR, and a Sensors and Software Noggin 500 GPR. Readings from the IMU magnetometer were excluded due to intermittent magnetic interference commonly found in indoor environments. Ground truth data was acquired by a Leica TS15 total station. The base station logs measurements from the onboard computer and the total station to ensure consistent timing.

In our testing, we found that GPR antenna placement was a critical design decision. Our initial design suspended the radar system 2.5cm above the ground to maintain a fixed transformation between the IMU and GPR sensor. While a fixed sensor setting like this is most common for localization systems, the small air gap between the GPR and the ground introduced ground energy losses and multi-path interactions causing poor depth penetration and repetitive ringing. To address this, we added a passive suspension to the GPR system so as to maintain constant ground contact. This improved depth penetration as shown in the right image of Fig.~\ref{fig:exp_setup}(c). We address methods to improve the robustness and applicability of our system in Section~\ref{sec:discussion}.

\begin{figure*}[!t]
    \centering
    \includegraphics{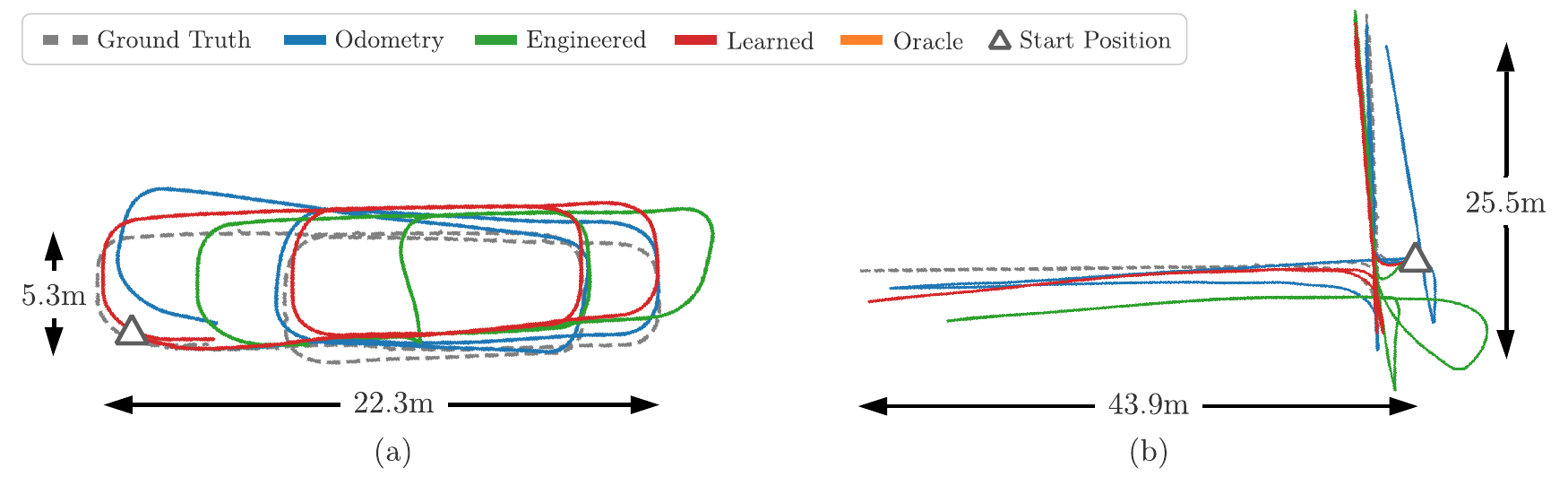}
    \caption{Qualitative evaluation of vehicle trajectory estimate over time.  \textbf{(a)} The robot follows a trajectory that loops over itself twice.  \textbf{(b)} The robot starts moving on the vertical segment, then moves along the horizontal segment, before finally revisiting the vertical segment. In both cases, the learned sensor model follows the ground truth data better than the engineered method despite receiving the same loop closure detections.}
    \label{fig:results:traj:qual}
\end{figure*}

\begin{figure*}[!t]
    \centering
    \includegraphics{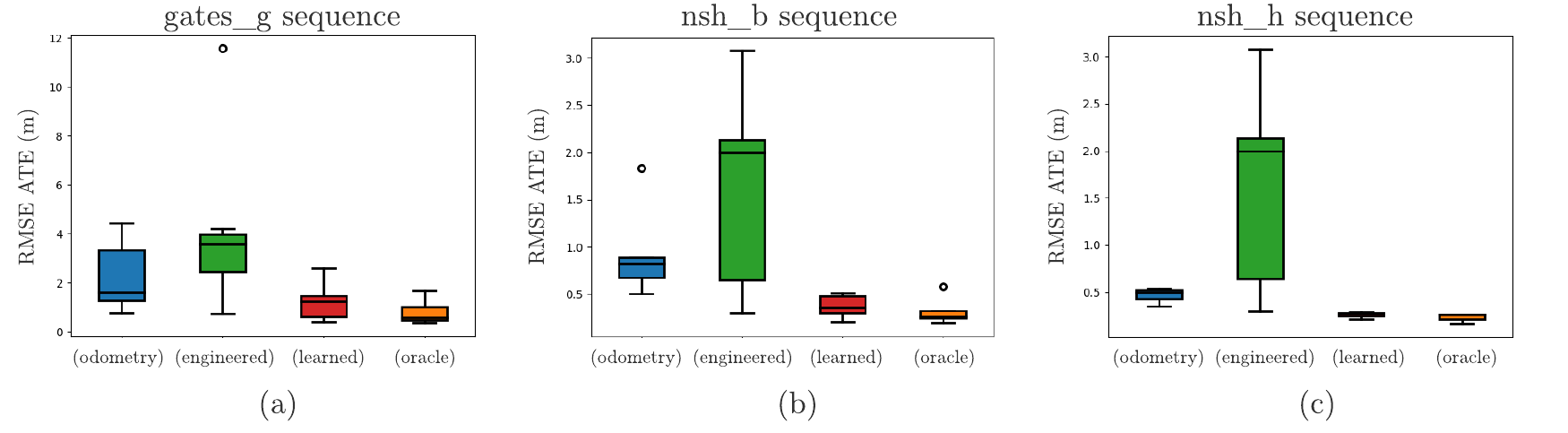}
    \caption{RMSE Absolute Trajectory Error (ATE) for each testing environment. }
    \label{fig:results:ate_per_set}\vspace{-4mm}
\end{figure*}

\begin{figure}
    \centering
    \includegraphics{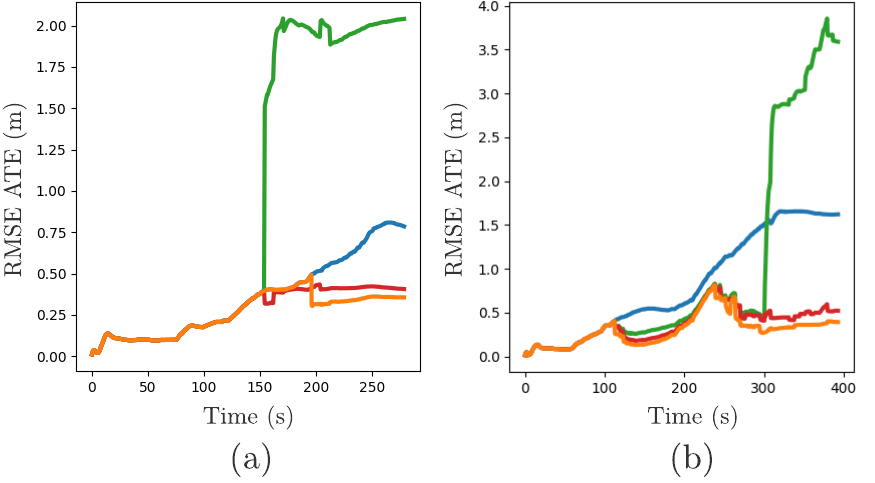}
    \caption{RMSE ATE over time. \textbf{(a)} Corresponds to the trajectory from Fig.~\ref{fig:results:traj:qual}(a). \textbf{(b)} Corresponds to the trajectory from Fig.~\ref{fig:results:traj:qual}(b). In both cases, loop closure events reduce the accumulation of odometry error.}
    \label{fig:results:ate_over_time}\vspace{-7mm}
\end{figure}

\subsection{Engineered GPR model baseline}
\label{sec:approach:analytic}

For evaluating our approach, we also consider a baseline \textit{engineered} GPR model that makes use of a correlation metric similar to existing work. This \textit{engineered} GPR model is incorporated as factors in the graph in the same way as our learned GPR model.

Existing GPR-based localization methods use a measurement correlation value to register images to a prior map~\cite{Cornick2016, Ort2020}. We compute a similar measurement correlation value between image pairs and select the pair with the highest correlation to add as factors in the graph. We use the Pearson product-moment correlation between two non-sequential GPR images $S_{t-k}, S_{t}$,
\begin{equation}
\label{eq:approach:analytic:corr}
    r(A(T), B(T)) = \frac{\sum_{r,c}{A_{r,c}(T) B_{r,c}(T)}}{\sqrt{\sum_{r,c} A_{r,c}(T)^2}\sqrt{\sum_{r,c} B_{r,c}(T)^2}}
\end{equation}
where, $A_{r,c}(T) = S_{t-k}(T, r, c) - \operatorname{mean}(S_{t-k}(T))$, $B_{r,c}(T) = S_t(T, r,c) - \operatorname{mean}(S_t(T))$, $r$ is the submap row, $c$ is the submap column, and $T$ is the transformation from $S_{t-k}$ to $S_t$ used to only compare the shared region between the submaps. 

Specifically, we estimate the transformation along the x-direction of the robot's motion. This is obtained by solving the optimization that maximizes the correlation from Eq.~\ref{eq:approach:analytic:corr},
\begin{equation}
\label{eq:approach:analytic:opt}
    z_{t-k,t} = K\ \underset{{T}}{\operatorname{argmax}} \left( r(A(T), B(T)) \right)
\end{equation}
where, $K$ is a constant that converts submap pixel space to robot motion space.

\subsection{GPR factor learning}
\label{sec:eval:learning}
We now evaluate the performance of different GPR models to be used in the factor graph. Table~\ref{tab:eval:tform_loss} compares losses for the baseline \textit{engineered} models against our learned GPR sensor model discussed in Section~\ref{sec:approach:senor_model} for different choices of network architectures. The loss here is the mean-squared error that we saw in Eq. \ref{eq:approach:gpr:eq1} against ground truth transformations. We use the Huber loss function for robustness to outliers. 
\textit{engineered} is the baseline correlation approach discussed in Section~\ref{sec:approach:analytic} which resembles prior work~\cite{Cornick2016}. \textit{zeroth} is a zeroth-order model that predicts the average relative transform of the training dataset. \textit{corr-feat} are acquired by concatenating the $\operatorname{argmax}$ values of the discrete cost curves from the spatial correlation of autoencoder activation maps as described in Section~\ref{sec:approach:senor_model}. \textit{linear} and \textit{nonlinear} refer to the linear and nonlinear activation functions in the fully-connected layers.

The \textit{engineered} model has notably high losses as there can be many false positive matches since the GPR image data often has similar amplitude with subtle features. In comparison, we see that the different learned GPR model architectures have much lower losses, with \textit{linear, corr-feat} model type having the best performance. We also found that spatial correlation features \textit{corr-feat} generalized much better than using vectorized autoencoder feature maps directly.


\begin{table}[!t]
	\small
    \centering
    \caption{\small Validation losses for different GPR models (cm)}
    \resizebox{\columnwidth}{!}{
	\begin{tabulary}{\columnwidth}{LCCCC}\toprule
    & \multicolumn{4}{c}{ {\bf Dataset} } \\
    \bf{Model type} & {gates\_g} & {nsh\_b} & {nsh\_h} & Combined \\ \midrule
	engineered & 97.66 & 88.07 & 100.24 & 95.33 \\
	zeroth & 17.81 & 137.97 & 23.30 & 59.49  \\
	\bf{linear, corr-feat}  & \bf{14.81} & \bf{2.88} & \bf{3.90} & \bf{7.40}\\
	nonlinear, corr-feat & 14.96 & 3.09 & 4.06 & 8.97  \\
    \bottomrule
    \end{tabulary}
    }
    \label{tab:eval:tform_loss}\vspace{-4mm}
\end{table}

\subsection{Factor graph optimization}
\label{sec:eval:opt}

We finally demonstrate the effect of the GPR sensor model on odometry drift correction through a qualitative and quantitative evaluation. The best performing model from Table~\ref{tab:eval:tform_loss}, \textit{linear, corr-feat}, was selected as the \textit{learned} model in this evaluation. We compare the \textit{engineered} and \textit{learned} model with the ground truth trajectory and the results from an \textit{oracle} model. The \textit{oracle} model predicts the ground truth relative pose transformation, representing the lower error bound of the GPR factor. The covariance parameters in the graph are fixed and the same across sequences.

\textbf{Qualitative Evaluation: } Both trajectories in Fig.~\ref{fig:results:traj:qual} were collected in a modern parking garage with homogeneous and repetitious subsurface features, making the test challenging for our GPR system. Additionally, non-deterministic reflections from nearby cars and walls were present in the dataset. As shown, the sensor model relying on \textit{engineered} features identifies incorrect transformations by comparing processed submaps directly, causing substantial drift. This is also reflected in Fig.~\ref{fig:results:ate_over_time}, where incorrect loop closures cause the \textit{engineered} model to have greater error than odometry alone. We see that the \textit{learned} model recovers the robot's poses close to the true trajectory and matches the \textit{oracle} performance closely. 


\textbf{Quantitative Evaluation: } Fig.~\ref{fig:results:ate_per_set} shows the RMSE of the absolute trajectory error described in~\cite{sturm12iros}. Errors were computed over 7 sequences in \textit{gates\_g}, 5 sequences in \textit{nsh\_b}, and 3 sequences in \textit{nsh\_h} (15 sequences overall). Odometry measurements had greater error in the \textit{gates\_g} set since loops and turns were more common. To better compare sensor models, an oracle provided identical loop closure observations to both \textit{engineered} and \textit{learned} models. The \textit{learned} model outperformed the \textit{engineered} model in all sequences. The \textit{engineered} model would often produce false detections in sparsely featured environments, causing a large variance in performance. Repetitive structures and noise in the processed image caused the \textit{engineered} model to perform poorly. The \textit{learned} sensor model performed consistently better by decomposing the image into feature activation maps, containing prominent structures and edges in the data that are easier to compare. The \textit{learned} model nearly performs as well as the \textit{oracle} detector, the model's lower error bound.

\section{Discussion}
\label{sec:discussion}

We presented a GPR-based localization system that uses a learned sensor model to position a robot in a unknown, GPS-denied environment. We accomplished this by combining GPR measurements to create individual submaps, which were provided to a transform prediction network to predict a relative pose transformation. These transformations were then incorporated as measurement likelihoods in a factor graph, where the GPR measurement corrected for accumulated drift from proprioceptive measurements. We validated our system in three distinct environments, where we have shown improved localization performance over existing correlation-based approaches. 

Prior work has relied on sophisticated multi-channel array GPR systems to perform localization. In this evaluation, we demonstrate that a low-cost, off-the-shelf, single-channel GPR system can be used for localization. While our evaluation was performed with a ground-coupled GPR system, our work can generalize for air-launched GPR systems, improving the viability for real world deployment.

Our learned sensor model is trained using data collected at a single point in time and is not currently equipped for long-term prior map registration, since measurements are affected by changes in subsurface moisture content. In the future, we would like to extend this work by using data collected from highly accurate, multi-channel GPR systems during different seasons and weather conditions to improve the robustness of prior map registration.

Current GPR-based localization systems must revisit explored locations to correct for accumulated drift. Future work will investigate methods to attenuate this requirement by modeling the typical relationships of continuous subsurface structures (eg. pipes, subsurface mines, natural caves), enabling more robust GPR-based state estimation in a broader range of settings where revisitation is not practical. We hope to build on this work and move towards general representations for GPR data needed to enable robust, real world deployment of localizing GPR systems.

\section{Acknowledgements}
\footnotesize{
The authors are grateful to Dimitrious Apostolopolous and Herman Herman for providing the Noggin 500 GPR and Sebastian Scherer for providing the XSENS IMU used to support this study. We thank David Kohanbash for operating the total station used to acquire ground truth positions for this study, Robert Bittner for helping us use the National Robotics Engineering Center facilities for testing, and Ryan Darnley for supporting the initial development of the baseline inertial navigation system.}

\balance

\footnotesize
\bibliographystyle{ieeetr}
\bibliography{references.bib}

\end{document}